\documentclass[runningheads]{llncs}

\usepackage[T1]{fontenc}
\usepackage{graphicx}
\usepackage{amsmath,amssymb}
\usepackage{algorithm}
\usepackage{algpseudocode}
\usepackage{booktabs}
\usepackage{multirow}
\usepackage{array}
\usepackage{xcolor}

\begin{document}

\title{MARIC: Multi-Agent Reasoning for Image Classification}
\titlerunning{MARIC}

\author{Wonduk Seo\inst{1,2}\thanks{Equal Contribution.}\orcidID{0009-0008-6070-1833} \and
Minhyeong Yu\inst{1}\textsuperscript{*}\orcidID{0000-0002-0350-4349} \and
Hyunjin An\inst{1}\orcidID{0009-0005-5102-9983} \and
Seunghyun Lee\inst{1}\orcidID{0009-0000-1687-1597}
}
\authorrunning{Seo et al.}

\institute{Enhans, Seoul, South Korea\\
\email{\{minhyeong, hyunjin, seunghyun\}@enhans.ai}
\and
Peking University, Beijing, China\\
\email{seowonduk@pku.edu.cn}
}

\maketitle

\begin{abstract}
Image classification has traditionally relied on dataset-specific training and fine-tuning, while recent vision-language models (VLMs) enable training-free prediction via prompting. However, single-pass VLM inference often misses complementary visual evidence (e.g., object attributes vs. context cues), yielding brittle or under-justified decisions. We introduce \textbf{Multi-Agent Reasoning for Image Classification (MARIC)}, an inference-time framework that decomposes classification into (i) an \emph{Outliner Agent} that summarizes the global theme and generates orthogonal aspect prompts, (ii) $K$ \emph{Aspect Agents} that extract complementary evidence under these prompts, and (iii) a \emph{Reasoning Agent} that performs an explicit reflection step to resolve conflicts and synthesize a unified prediction with an interpretable reasoning trace.
In the default setting, MARIC uses sequential calls ($K{=}3$), balancing diversity and redundancy. Experiments on four image classification benchmarks demonstrate that MARIC outperforms standard prompting baselines, highlighting the effectiveness of multi-agent visual reasoning for robust and interpretable image classification.
\end{abstract}
\section{Introduction}
Image classification is a cornerstone of modern computer vision, underpinning applications ranging from content moderation to medical decision support. For more than a decade, progress has largely been driven by supervised deep learning, where models learn discriminative representations from large-scale annotated datasets~\cite{lecun1989backpropagation,deng2009imagenet}. Convolutional neural networks (CNNs) established strong performance through hierarchical feature learning~\cite{scherer2010evaluation,krizhevsky2012imagenet}, and transformer-based architectures further improved accuracy and scalability by modeling global interactions~\cite{parmar2018image,dosovitskiy2020image}. Despite their success, these pipelines remain expensive to adapt: they typically require dataset-specific training (or extensive fine-tuning), careful hyperparameter tuning, and their predictions are often difficult to justify with explicit, human-readable evidence.

Recently, vision-language models (VLMs) have offered an appealing alternative by aligning visual inputs with natural language, enabling \emph{training-free} classification via prompting and label descriptions~\cite{radford2021learning,zhou2022learning}. In principle, this allows practitioners to deploy a single pretrained model across many label sets without collecting additional annotations. In practice, however, standard VLM inference is typically performed in a single pass, and the resulting prediction can be brittle: the model may latch onto a dominant cue (e.g., background context) while ignoring subtle but label-critical details (e.g., fine-grained texture or small objects). This issue becomes especially pronounced under distribution shifts or for confusable categories, where complementary evidence is needed to make reliable decisions. Moreover, even when a VLM produces a correct label, the accompanying explanation can be shallow or inconsistent with the actual visual evidence, limiting trust and debuggability~\cite{zhang2024visually}.

This paper explores a simple but effective premise: if we allocate additional computation \emph{at inference time}---without updating any model parameters---we can systematically collect diverse evidence and perform explicit cross-checking before committing to a final prediction. Motivated by recent progress in agentic reasoning, we propose \textbf{Multi-Agent Reasoning for Image Classification (MARIC)}, a collaborative inference framework that decomposes classification into complementary roles. First, an \emph{Outliner Agent} identifies the global theme of the image and generates a small set of targeted prompts designed to be minimally overlapping. \emph{Aspect Agents} subsequently query the VLM under these prompts to extract complementary evidence (e.g., object attributes, contextual cues, and discriminative details). Finally, a \emph{Reasoning Agent} aggregates the evidence, explicitly reflects on agreement and conflicts among aspects, and produces both the predicted label and an interpretable reasoning trace.

By structuring VLM inference as evidence gathering followed by reflective fusion, MARIC aims to improve both robustness and transparency: failures become easier to diagnose because the decision is grounded in explicit aspects, and the framework can down-weight unsupported claims when agents disagree. Extensive experiments on four diverse benchmarks show that this decomposition yields consistent improvements over standard prompting and reasoning-oriented baselines.

\paragraph{Contributions.}
\begin{itemize}
  \item We introduce \textbf{MARIC}, an inference-time multi-agent framework for VLM-based image classification that outputs both a predicted label and an interpretable reasoning trace.
  \item We formalize \emph{orthogonal aspect prompting}, where an Outliner Agent generates targeted, minimally-overlapping aspect prompts to encourage complementary evidence extraction across Aspect Agents.
  \item We empirically demonstrate that reflective evidence fusion improves robustness across diverse datasets, highlighting the value of multi-agent decomposition for training-free classification.
\end{itemize}

\begin{figure}
  \centering
  \includegraphics[width=0.8\linewidth]{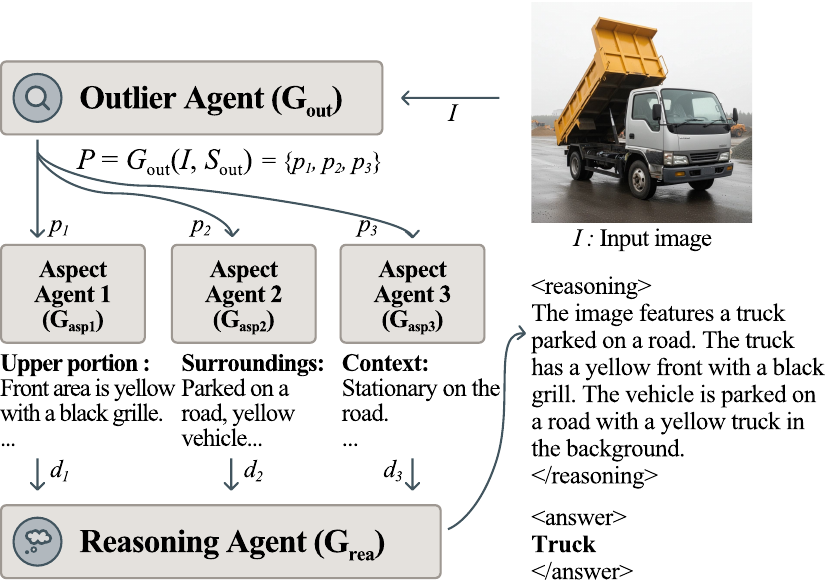}
  \caption{High-level overview of MARIC. The Outliner Agent proposes aspect prompts, Aspect Agents extract complementary evidence, and the Reasoning Agent reflects to produce the final prediction and trace.}
  \label{fig:maric-overview}
\end{figure}

\section{Related Work}

\subsection{Vision-Language Foundation Models}
Image classification has traditionally relied on parameter-intensive model training, requiring large annotated datasets and extensive fine-tuning to achieve competitive performance~\cite{lecun1989backpropagation,deng2009imagenet,scherer2010evaluation}. To mitigate these constraints, early vision-language research focused on joint embedding learning for tasks such as image captioning and image-text retrieval. The combination of large-scale image-text pretraining and transformer architectures further accelerated progress~\cite{radford2021learning,zhou2022learning}. Representative models such as CLIP and Flamingo learned robust joint representations that narrow the semantic gap between vision and language~\cite{radford2021learning,alayrac2022flamingo}, demonstrating strong zero-shot and few-shot generalization across diverse tasks including image captioning, visual question answering (VQA), text-to-image synthesis, and cross-modal retrieval. Nevertheless, these models still rely on single-pass representations and static alignment, often failing to capture complementary visual cues, motivating subsequent work on adaptive zero-shot classification and reasoning-enhanced methods.

\subsection{Zero-Shot VLM Classification}
Zero-shot classification, where class names are embedded as textual prompts and predictions are made via image-text similarity~\cite{radford2021learning}, has become a standard paradigm. Accuracy has been improved through prompt tuning, adapter-based fine-tuning, and alongside multimodal prompt alignment~\cite{abdul2023align}, ensemble prompting, and reasoning-driven approaches~\cite{masoud2024exploring}. More recently, role-differentiated multi-agent frameworks iteratively propose, critique, and refine candidate solutions to achieve more robust and interpretable classification~\cite{seo2025vispath}. Nonetheless, many approaches still rely on single-pass inference, limiting their ability to capture complementary cues and produce transparent reasoning. This motivates frameworks such as \textbf{MARIC}, which explicitly decomposes image classification into Outliner, aspect, and reasoning agents, generating complementary evidence and structured reasoning traces that enhance both accuracy and interpretability.

\section{Method}
\subsection{Problem setup and overview}
We consider zero-shot image classification in which a pretrained VLM is used \emph{without} any parameter updates. Given an input image $I$ and a set of candidate labels $\mathcal{Y}$ (optionally with natural-language label descriptions), the goal is to predict a label $\hat{y}\in\mathcal{Y}$ and provide a concise, human-readable justification.

MARIC reframes this task as \emph{inference-time evidence collection and reflective fusion}. Instead of relying on a single prompt and a single model response, MARIC decomposes inference into three roles (Fig.~\ref{fig:maric-overview}):
(i) an \emph{Outliner Agent} that proposes complementary aspect prompts, (ii) $K$ \emph{Aspect Agents} that extract evidence under these prompts, and (iii) a \emph{Reasoning Agent} that reconciles the evidence via an explicit reflection step before emitting the final prediction.

\subsection{Outliner Agent $G_{\text{out}}$ (orthogonal aspect prompting)}
The Outliner Agent produces a small set of prompts $\mathcal{P}=\{p_1,\dots,p_K\}$ conditioned on the image and label set:
\begin{equation}
  \mathcal{P} = G_{\text{out}}(I,\mathcal{Y}\mid S_{\text{out}}).
\end{equation}
Each prompt $p_k$ is designed to elicit an \emph{evidence type} rather than a final label. In practice, we use a prefix--postfix structure
that specifies (a) \emph{where/what to attend to} (prefix) and (b) \emph{what to report} (postfix). The key requirement is \emph{orthogonality}: prompts should cover complementary visual cues with minimal redundancy. Concretely, our default setting uses $K{=}3$ role-typed aspects:
\textbf{(1) object-centric} cues (shape/texture/color/parts), \textbf{(2) context-centric} cues (scene, background, co-occurring objects), and \textbf{(3) discriminative} cues (rare details that separate confusable labels).

\subsection{Aspect Agents $G_{\text{asp}}$ (evidence extraction)}
Given prompts $\mathcal{P}$, MARIC runs $K$ independent Aspect Agents as separate VLM calls. For each $p_k\in\mathcal{P}$, the Aspect Agent returns a description $d_k$:
\begin{equation}
  d_k = G_{\text{asp}}(I, p_k\mid S_{\text{asp}}).
\end{equation}
We instruct Aspect Agents to (i) focus on concrete visual facts, (ii) avoid generic captions, and (iii) avoid repeating information already covered by other aspects (as signaled by the Outliner prompts). This produces a compact evidence set $D=\{d_1,\dots,d_K\}$ that is both diverse and traceable to specific cues.

\subsection{Reasoning Agent $G_{\text{rea}}$ (reflective fusion)}
The Reasoning Agent aggregates the evidence set $D$ and produces a final label $\hat{y}$ together with a reasoning trace $r$:
\begin{equation}
  (\hat{y}, r) = G_{\text{rea}}(I,\mathcal{Y}, D\mid S_{\text{rea}}).
\end{equation}
Crucially, MARIC includes an explicit \emph{reflection} step before the final decision. During reflection, $G_{\text{rea}}$ (i) checks agreement and conflicts among $\{d_k\}$, (ii) flags unsupported or speculative claims, and (iii) prioritizes evidence that is most discriminative for labels in $\mathcal{Y}$. The output trace is structured to cite the aspects that most strongly support the chosen label, improving transparency and debuggability.

\subsection{Inference procedure}
Algorithm~\ref{alg:maric} summarizes the full inference-time pipeline.

\begin{algorithm}
\caption{MARIC inference (sketch)}
\label{alg:maric}
\begin{algorithmic}[1]
\Require Image $I$, candidate labels $\mathcal{Y}$, number of aspects $K$
\State $\{p_k\}_{k=1}^K \leftarrow \textsc{Outliner}(I, \mathcal{Y})$
\For{$k=1$ to $K$}
  \State $d_k \leftarrow \textsc{AspectAgent}(I, p_k)$
\EndFor
\State $\hat{y}, r \leftarrow \textsc{Reasoner}(I, \mathcal{Y}, \{d_k\}_{k=1}^K)$
\State \Return $\hat{y}$, reasoning trace $r$
\end{algorithmic}
\end{algorithm}

\subsection*{Use of AI Tools}
A large language model was used to assist with improving the readability, wording, and organization of the manuscript. The authors reviewed and revised all AI-assisted outputs and take full responsibility for the accuracy and integrity of the final manuscript.

\section{Experiments}

\subsection{Setup}

\subsubsection{Experimental Datasets}
We evaluate \textbf{MARIC} on four image benchmark datasets:

\textbf{CIFAR-10}~\cite{krizhevsky2009learning}: A canonical 10-class benchmark (airplane, automobile, bird, cat, deer, dog, frog, horse, ship, truck), sampled 100 images per class.

\textbf{OOD-CV}~\cite{zhao2022ood}: An out-of-distribution robustness benchmark with 10 classes (aeroplane, bicycle, boat, bus, car, chair, diningtable, motorbike, sofa, train), sampled 100 images per class.

\textbf{Weather Dataset}~\cite{ajayi_srivastava_2023_weather}: Weather condition classification with 1,125 images across 4 classes (sunrise, shine, rain, cloudy).

\textbf{Skin Cancer Dataset}~\cite{venugopal_joseph_das_nath_2023_skin_cancer}: Binary melanoma detection from DermIS and DermQuest with 87 balanced images per class (healthy/cancerous).

We emphasize that our goal is \emph{training-free} evaluation of inference-time reasoning rather than data-scaling effects; thus, we follow a controlled sampling protocol to keep evaluation balanced across classes and datasets.

\subsubsection{Models Used}
To evaluate the effectiveness of \textbf{MARIC}, we adopt $2$ representative Vision-Language Models (VLMs), including: (1) llava-1.5-7b-hf model, and (2) llava-1.5-13b-hf~\cite{liu2024improved}. These models were selected to represent a spectrum of parameter scales and architectural designs, and configured with a temperature of $0$ to ensure precise and focused outputs.

\subsubsection{Baseline Methods}
We compare \textbf{MARIC} against $3$ representative baselines, including: (1) \emph{Direct Generation}, where the VLM produces the classification result directly from the input image without additional prompting or reasoning~\cite{zhang2024visually}; (2) \emph{Chain-of-Thought (CoT)}, in which the model is explicitly instructed to reason step-by-step~\cite{wei2022chain}; (3) \emph{Single-Agent Visual Reasoning (SAVR)} baseline, where a single handcrafted prompt guides the model to generate both reasoning and classification in one pass.

\subsection{Experiment Results}

\begin{table*}[t]
\centering
\caption{Classification accuracy (\%) on benchmark datasets. All VLM-based methods are evaluated in a \emph{zero-shot, no-parameter-update} setting using the same backbone per block (LLaVA 1.5-7B / 13B). Best results are in \textbf{bold}, second-best are \underline{underlined}. We will additionally report inference cost (tokens/latency) in the revised version.}
\label{tab:main_results}
\resizebox{0.9\textwidth}{!}{%
\begin{tabular}{llccccc}
\toprule
\textbf{Model} & \textbf{Method} & \textbf{CIFAR-10} & \textbf{OOD-CV} & \textbf{Weather} & \textbf{Skin Cancer} & \textbf{Avg. Score} \\
\midrule
\multirow{4}{*}{LLaVA 1.5-7B}
    & Direct Generation & 64.8  & 66.5  & 50.2  & \underline{50.6}  & 58.03 \\
    & Chain-of-Thought (CoT) & \underline{83.5}  & \underline{80.8}  & \textbf{70.1}  & 50.0  & 71.10 \\
    & SAVR & 75.6  & 55.5  & 48.0  & 49.4  & 57.13 \\
    & \textbf{MARIC (Ours)} & \textbf{90.8} & \textbf{81.9} & \underline{65.6} & \textbf{50.6} & \textbf{72.23} \\
\midrule
\multirow{4}{*}{LLaVA 1.5-13B}
    & Direct Generation & 86.6  & \underline{86.2}  & 21.7  & 52.9  & 61.85 \\
    & Chain-of-Thought (CoT) & 88.0  & 75.2  & \underline{81.1} & 49.4 & 73.43 \\
    & SAVR & \underline{88.6}  & 81.2  & 63.0  & \textbf{62.6}  & \underline{73.85} \\
    & \textbf{MARIC (Ours)} & \textbf{93.5} & \textbf{89.9} & \textbf{85.2} & \underline{56.3} & \textbf{81.23} \\
\bottomrule
\end{tabular}%
}
\end{table*}

\subsubsection{Main Results}
Across all $4$ benchmark datasets (see Table~\ref{tab:main_results}), baseline methods show clear strengths but also notable shortcomings. \emph{Direct Generation}, while simple and computationally efficient, often produces shallow predictions that overlook complementary cues in the image. \emph{Chain-of-Thought (CoT)} provides step-by-step reasoning, yet its additional verbal reasoning does not always translate into higher accuracy. The \emph{Single-Agent Visual Reasoning (SAVR)} baseline captures reasoning and classification in one pass, but its reliance on a single handcrafted prompt limits evidence diversity and can cause the model to miss subtle attributes that are important for robust decisions.

\textbf{MARIC} achieves the highest average score for both VLM backbones. With LLaVA 1.5-7B, \textbf{MARIC} obtains an average accuracy of $72.23\%$, improving over Direct Generation by $14.20$ points and over SAVR by $15.10$ points, while also slightly outperforming CoT by $1.13$ points. The improvement is especially clear on CIFAR-10, where \textbf{MARIC} reaches $90.8\%$ accuracy, exceeding the strongest baseline by $7.3$ points. On OOD-CV, \textbf{MARIC} also produces the best result ($81.9\%$), suggesting that decomposed visual evidence is useful when the test distribution differs from standard in-distribution images. Weather classification is the main exception for the 7B model, where CoT performs better ($70.1\%$ vs. $65.6\%$), indicating that a smaller VLM can still benefit from direct step-by-step reasoning when the label space is compact and defined by global scene conditions.

The gains become more pronounced with LLaVA 1.5-13B. \textbf{MARIC} reaches an average accuracy of $81.23\%$, outperforming the second-best method, SAVR, by $7.38$ points. It achieves the best result on CIFAR-10 ($93.5\%$), OOD-CV ($89.9\%$), and Weather ($85.2\%$), with margins of $4.9$, $3.7$, and $4.1$ points over the strongest corresponding baselines. These results suggest that the larger backbone can better exploit the aspect-level evidence generated by \textbf{MARIC}, because the Reasoning Agent has sufficient capacity to compare complementary descriptions and suppress unsupported cues. On Skin Cancer, however, SAVR achieves the highest accuracy ($62.6\%$), while \textbf{MARIC} obtains the second-best result ($56.3\%$). This gap suggests that medical binary classification may require more domain-specific prompting or expert-oriented visual criteria than the general-purpose aspect prompts used in the current framework.

Overall, the results indicate that the benefit of \textbf{MARIC} is not simply due to longer outputs or more inference steps. Instead, the role decomposition encourages the model to gather object-centric, context-centric, and discriminative evidence before making a prediction. This structured evidence collection is particularly helpful for multi-class natural-image datasets and robustness-oriented evaluation, where multiple visual cues must be balanced. Beyond accuracy, \textbf{MARIC} also improves interpretability: its final predictions are accompanied by explicit aspect descriptions and a reflective synthesis, making it easier to inspect which cues supported the decision and where possible conflicts arose.

\begin{table}[t]
\centering
\caption{Ablation study on \textbf{MARIC} components using LLaVA 1.5-13B.}
\label{tab:ablation_components}
\resizebox{0.7\textwidth}{!}{%
\begin{tabular}{lcccc}
\toprule
\textbf{Configuration} & \textbf{CIFAR-10} & \textbf{OOD-CV} & \textbf{Weather} & \textbf{Skin Cancer} \\
\midrule
Full \textbf{MARIC} & \textbf{93.5} & \textbf{89.9} & \textbf{85.2} & \textbf{56.3} \\
w/o Aspect Agents & 93.4 & 89.4 & 84.5 & 52.9 \\
\bottomrule
\end{tabular}%
}
\end{table}

\subsection{Ablation Study on Agent Components}
To evaluate the contribution of aspect agents in \textbf{MARIC}, we conducted an ablation study using the LLaVA 1.5-13B model~\cite{liu2024improved} (Table~\ref{tab:ablation_components}). Results show that the full framework achieves the best performance across all $4$ benchmarks, with notable gains on Weather and Skin Cancer classification. Even when the \emph{Aspect Agents} are removed, \textbf{MARIC} still maintains strong accuracy, indicating that global prompts and reflective reasoning alone provide competitive performance.

\begin{figure}[t]
    \centering
    \includegraphics[width=0.8\linewidth]{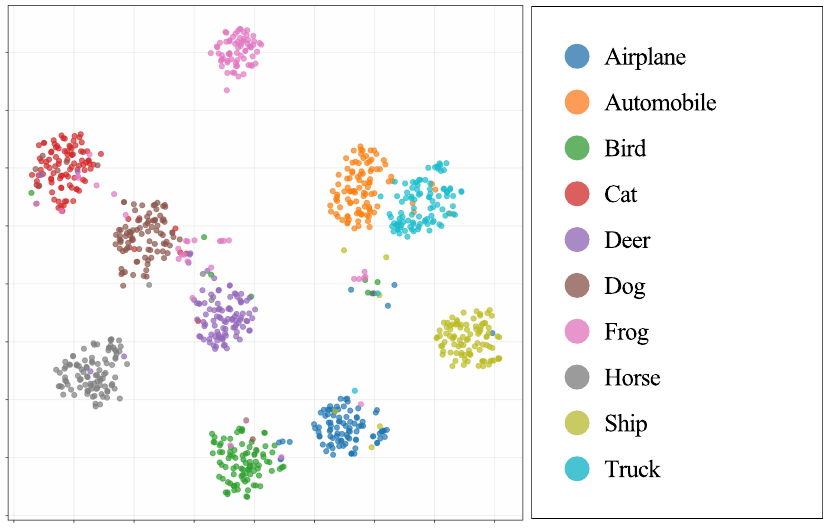}
    \caption{t-SNE visualization of reasoning embeddings generated by \textbf{MARIC} on CIFAR-10. Each class forms a well-separated and compact cluster, contributing to both robustness and interpretability.}
    \label{fig:tsne}
\end{figure}

Additionally, Figure~\ref{fig:tsne} presents a t-SNE visualization of reasoning text embeddings generated by \textbf{MARIC} for CIFAR-10 classification. Each point corresponds to an image sample encoded with E5~\cite{wang2022text}, colored by its true class. The clusters reveal meaningful relationships: animal classes are separated from vehicle classes, while bird and airplane clusters appear close, reflecting their shared ``sky/flying'' semantics. This indicates that \textbf{MARIC}'s reasoning captures nuanced distinctions beyond categorical boundaries.

\begin{table}[t]
\centering
\caption{Qualitative Study Results (M $\pm$ SD).}
\label{tab:qualitativestudy}
\resizebox{0.8\textwidth}{!}{%
\begin{tabular}{ccc}
\toprule
\textbf{Aspect Relevance} & \textbf{Aspect Diversity} & \textbf{Description Accuracy} \\
\midrule
3.93 $\pm$ 1.08 & 3.97 $\pm$ 1.07 & 4.00 $\pm$ 1.05 \\
\bottomrule
\end{tabular}%
}
\end{table}

\section{Qualitative Analysis}
To further evaluate the quality of \textbf{MARIC}'s aspect decomposition, we conducted a human study on 30 randomly sampled images from CIFAR-10, chosen for its multi-class nature. Eleven participants (AI researchers and data scientists in their 30s, independent from this study; M = 30.64, SD = 2.54) evaluated the three aspects generated by the Aspect Agent and their corresponding descriptions. Each set of aspects was rated on three criteria - Aspect Relevance, Aspect Diversity, and Description Accuracy - all using a 5-point Likert scale. Results show a mean relevance score of 3.93, diversity score of 3.97, and accuracy score of 4, indicating that the Aspect Agent generates meaningful, complementary aspects and faithfully describes them (see Table~\ref{tab:qualitativestudy}). These results suggest that the aspects help capture distinctive characteristics and complement each other, while descriptions remain faithful to the visual content. In particular, the high diversity score supports the intended role of orthogonal aspect prompting: agents tend to attend to different visual evidence rather than producing redundant captions. The strong description accuracy score further indicates that the resulting reasoning traces are not only informative, but also grounded in observable image content.

\section{Conclusion}
In this paper, we introduced \textbf{MARIC}, a novel multi-agent framework for image classification that combines Multi-Agent into a collaborative reasoning process. By generating targeted prompts, extracting complementary fine-grained descriptions, and reflecting on them before synthesis, \textbf{MARIC} produces robust and interpretable classification decisions. Extensive experiments on four diverse benchmark datasets confirm that this agent-based collaboration leads to consistent and significant improvements in accuracy.

\section{Limitation and Future Work}
While MARIC improves zero-shot image classification via coordinated \emph{Outliner}, \emph{Aspect}, and \emph{Reasoning} agents, it has several limitations.
First, MARIC incurs higher inference overhead due to multi-step prompting ($1{+}K{+}1$ calls), which increases latency and token usage compared to single-pass baselines.
Second, performance is sensitive to prompt quality: if the Outliner Agent produces overlapping or mis-specified aspect prompts, Aspect evidence may become redundant or omit critical cues. Third, we use a fixed default $K{=}3$ in the main experiments; although effective, this may not be optimal for all domains. Future work will focus on fine-tuning Reasoning agents to reduce redundancy and conducting larger, more detailed qualitative studies to isolate failure modes and enable adaptive agent scheduling. We also plan to explore cost-aware variants of MARIC that dynamically adjust the number of aspect agents according to image difficulty, confidence, or detected ambiguity. For domain-specific tasks such as medical image classification, future work should incorporate expert-designed aspect prompts and validation protocols to improve reliability while preserving the interpretability benefits of multi-agent reasoning.

\bibliographystyle{splncs04}
\bibliography{prev_refs}

\end{document}